\documentclass[twocolumn,a4paper,10pt]{article}
\usepackage[a4paper,margin=1.8cm,columnsep=0.7cm]{geometry}
\usepackage{amsmath,amssymb,bm}
\usepackage{graphicx,booktabs,array}
\usepackage{caption,subcaption}
\usepackage[hidelinks]{hyperref}
\usepackage{enumitem}
\title{Fast Construction of Inner Sphere Trees Using Machine
Learning}
\author{Jianneng Huang and Wenjia Rao$^{*}$\\[3pt]
Department of Physics, Hangzhou Dianzi University, Hangzhou, China\\[2pt]
{\small $^{*}$Corresponding author: {wjrao@hdu.edu.cn}}}
\date{}

\begin{document}
\makeatletter
\twocolumn[{\begin{@twocolumnfalse}
\maketitle

\begin{abstract}
Collision detection for complex non-convex objects is a bottleneck in physics simulation, virtual reality, and robotics. Inner sphere trees (ISTs) answer proximity and penetration queries efficiently, but their greedy leaf-sphere construction is sequential and computationally expensive, which limits both large-scale preprocessing and runtime use. In this work, we present ML-IST, a machine-learning-based method that reformulates leaf-sphere generation as a single-shot set-prediction problem: a network predicts a fixed set of inscribed sphere candidates from multichannel voxel geometry, differentiable coverage feedback corrects the coarse prediction and fills residual gaps at finer scales, and a lightweight geometric projection enforces validity. Averaged over the test set, a single-GPU implementation of ML-IST reduces mean leaf-generation time by a factor of $2.0$--$2.6$ over single-core CPU construction across target capacities of 64--512 spheres, and qualitative results on complex models show multiscale sphere distributions that follow principal volumes and slender features. We further integrate ML-IST into the Jolt Physics and Bullet engines; in an alpha-puzzle benchmark it reaches 55.89~FPS, the highest among the compared representations (conventional IST, GImpact, and convex decomposition). These results support learned leaf generation as a practical component for fast, runtime construction of detailed collision representations.
\end{abstract}

\vspace{1em}
\end{@twocolumnfalse}}]
\makeatother

\section{Introduction}

Collision detection is essential to realism and responsiveness in virtual reality, physics simulation, robotic path planning, and medical interaction. For objects with complex non-convex features, such as holes, recesses, and thin structures, an effective collision representation must support efficient queries while accurately describing the occupied interior. Many acceleration structures have been developed for this purpose, including convex-body distance tests~\cite{gilbert1988}, convex-hull constructions~\cite{barber1996}, and exterior bounding-volume hierarchies~\cite{gottschalk1996,klosowski1998}. Sphere hierarchies approximate a model with nested spheres and have long been used for time-critical collision detection and medial-axis approximation~\cite{hubbard1996,bradshaw2004}. Among them, the inner sphere tree (IST)~\cite{weller2009a} places non-overlapping inscribed spheres inside the object and organizes them hierarchically, approximating the occupied volume compactly and supporting proximity and penetration queries.

Although ISTs are efficient at query time, their construction remains expensive. Conventional leaf generation uses a greedy iterative strategy over a signed distance field (SDF)~\cite{curless1996}, which encodes boundary clearance and inside--outside information and is widely used in collision-related computation~\cite{macklin2020}: the largest admissible inner sphere is repeatedly located in voxel space, and the residual feasible region is updated after every placement. The procedure is therefore inherently sequential and cannot fully exploit the parallelism of modern GPUs; construction time grows with voxel resolution, geometric complexity, and the target number of leaf spheres. Prior IST research has emphasized hierarchical organization, query performance, and penetration-depth estimation, whereas the cost of generating the leaf spheres has received less attention. Reducing this cost would benefit both large-scale preprocessing and rapid, runtime construction of collision representations.

Recent progress in deep learning has produced data-driven representations of three-dimensional geometry~\cite{wu2015,maturana2015} and neural implicit models such as DeepSDF and Occupancy Networks, which learn continuous geometric fields from spatial coordinates~\cite{park2019,mescheder2019}. For collision-related computation, Zesch et al.\ predict signed distances for deformable objects from a deformation code and a query point~\cite{zesch2022}, and neural builders have been used to accelerate the construction of spatial subdivisions such as k-d trees and BVHs~\cite{evangelou2023}. These methods demonstrate that search- or optimization-intensive geometric procedures can be transferred to offline training and approximated by a forward pass at runtime. They primarily predict query-level quantities, such as an SDF value, an occupancy probability, or a collision state, or they target exterior spatial subdivision for traversal optimization. ML-IST instead predicts an explicit, reusable set of interior spheres from multichannel voxel geometry, providing the leaves for a conventional IST hierarchy.

To this end, we propose ML-IST (machine-learning-based inner sphere tree generation), which reformulates leaf-sphere construction as the prediction of a fixed-capacity candidate set. Given a normalized 3D model, a multistage network predicts candidate centers, radii, and objectness scores from a multichannel voxel representation: it first establishes coarse coverage with large base spheres and then corrects them and fills residual gaps at finer scales through differentiable coverage feedback, after which a lightweight geometric projection enforces validity. Inference requires a single network forward pass. The contribution of this work is threefold. First, it formulates IST leaf generation as fixed-capacity sphere-set prediction. Second, it combines coarse prediction with differentiable coverage feedback and bounded multiscale refinement. Third, it evaluates generation time against the conventional greedy builder and demonstrates the predicted proxy in rigid-body and physics-engine benchmarks, including runtime construction inside Jolt Physics and Bullet.

The scope of this work is the leaf-sphere generation stage, which is the dominant construction cost in a conventional IST, rather than the replacement of the complete IST framework. The predicted spheres can still be organized into a conventional hierarchy and used by existing collision-query procedures, so ML-IST is a hybrid approach that moves much of the geometric search from online construction to offline training. This design supports both offline batch preprocessing and the runtime construction of detailed collision representations for objects that appear only after a physics engine has started; Section~\ref{sec:online_realtime} evaluates the latter setting.

\section{Learning-Based Inner Sphere Tree Generation}

Conventional IST leaf generation repeatedly searches for admissible inner spheres and updates the remaining volume. This sequential process becomes expensive as voxel resolution and target sphere count increase. ML-IST replaces the search with a learned mapping from voxel geometry to a fixed-capacity sphere set. The pipeline comprises multichannel input construction, base-sphere prediction, coverage-feedback generation, and multiscale refinement, followed by lightweight geometric projection.

\begin{figure*}[t]
    \centering
    \includegraphics[width=0.98\textwidth]{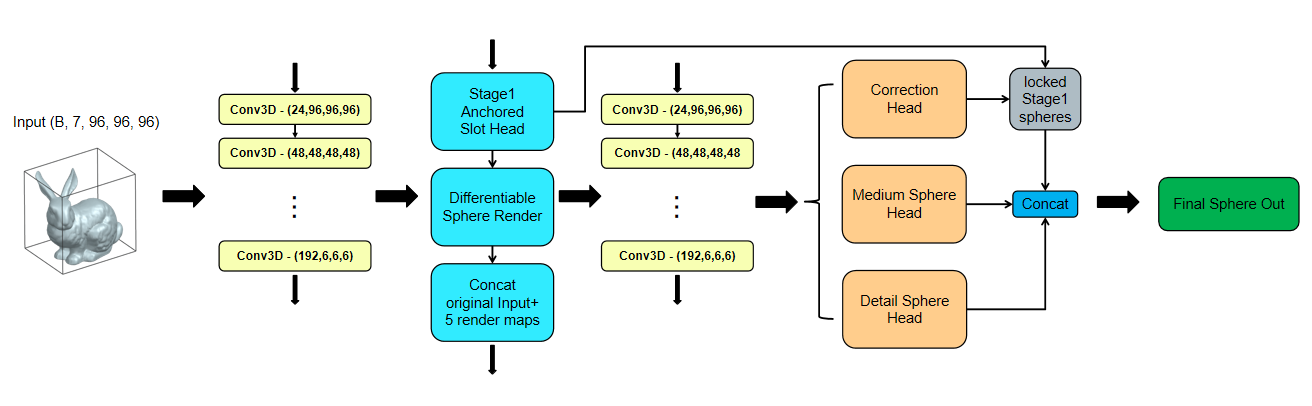}
    \caption{Overall ML-IST architecture. A seven-channel voxel representation is encoded into local and global geometric features for base-sphere prediction. A differentiable renderer generates five coverage-feedback maps, which are concatenated with the original input and processed by a second encoder. A correction head refines the base spheres, while medium- and detail-sphere heads predict additional candidates. The three groups are concatenated and geometrically projected to obtain IST leaves. Layer annotations give the output shape of each Conv3D stage as (channels, depth, height, width); for example, (24,96,96,96) denotes 24 channels at $96\times96\times96$ resolution, and $B$ is the batch size.}
    \label{fig:net_struct}
\end{figure*}

As illustrated in Fig.~\ref{fig:net_struct}, the network first predicts large base spheres to represent the principal interior regions. A differentiable renderer converts these spheres into voxel-level coverage feedback. A second encoder uses this feedback to guide limited corrections to the base spheres and predict medium and fine residual spheres for the remaining gaps. All branches are evaluated within a single network forward pass; the resulting leaves can subsequently be organized into a conventional IST hierarchy.

\subsection{Multichannel Voxel Geometry Representation}
\label{subsec:voxel_representation}

Let $\Omega_{\mathrm{in}}$ denote the interior of a 3D model. The objective is to generate a sphere set that covers the principal interior volume with limited exterior coverage and severe overlap under a finite candidate budget. ML-IST predicts $K$ candidates from the voxel input $\mathbf{V}$ in a single pass. Each candidate is represented by its center $\hat{\mathbf{c}}_i$, radius $\hat{r}_i$, and activation confidence, or objectness, $\hat{o}_i\in[0,1]$. Centers and radii are expressed in input-grid units, and the candidate set is denoted by $\hat{\mathcal{S}}$.

Model geometry is resampled onto a grid of resolution $R^3$ to construct the seven-channel input:
\begin{equation}
    \mathbf{V}=[M_{\mathrm{in}},D_{\mathrm{norm}},
    \Phi_{\mathrm{norm}},B,X,Y,Z].
    \label{eq:input_volume}
\end{equation}
The binary mask $M_{\mathrm{in}}$ equals 1 for interior voxels and 0 for exterior voxels. The interior distance field $D$ estimates, in grid units, the distance from an interior point to the model boundary $\partial\Omega$:
\begin{equation}
    D(\mathbf{p})\approx\min_{\mathbf{q}\in\partial\Omega}
    \|\mathbf{p}-\mathbf{q}\|_2,
    \qquad\mathbf{p}\in\Omega_{\mathrm{in}}.
    \label{eq:interior_distance}
\end{equation}
This distance approximates an upper bound on the admissible radius of a sphere centered at $\mathbf{p}$ and provides the geometric basis for subsequent constraints and radius projection. The implementation reads a preprocessed distance field when available; otherwise, it computes the Euclidean distance transform from the interior mask.

$D_{\mathrm{norm}}$ is scaled by the 99th percentile of interior distances, with a minimum scaling factor of 1, and clipped to $[0,1]$. The signed-distance channel $\Phi_{\mathrm{norm}}$ is robustly normalized by the 99th percentile of absolute SDF values and clipped to $[-1,1]$. The boundary channel $B=\exp(-6|\Phi_{\mathrm{norm}}|)$ emphasizes regions near the surface. Coordinate channels $X$, $Y$, and $Z$ normalize each axis to $[-1,1]$ and provide explicit spatial information~\cite{liu2018}.

The interior mask uses nearest-neighbor resampling, whereas continuous fields use trilinear interpolation. Normalized channels support feature extraction, while the unnormalized field $D$ is retained for sphere-validity evaluation so that normalized input values are not used directly as sphere radii.

\subsection{Gap-Aware Multistage Prediction}
\label{subsec:network}

The candidate capacity is divided into base, medium, and detail groups, with $K=K_0+K_m+K_d$. By default, the base and medium groups account for approximately $30\%$ and $32\%$ of the capacity, respectively, and the remaining slots are assigned to the detail group. These groups represent complementary scales of interior geometry rather than independent IST hierarchies.

\subsubsection{3D Residual Convolutional Encoder}

The first encoder extracts a local feature map $\mathbf{F}_0$ and a global shape descriptor $\mathbf{g}_0$ from $\mathbf{V}$. Each residual block contains two $3\times3\times3$ convolutions, group normalization~\cite{wu2018groupnorm}, SiLU activations, and a skip connection. Four spatial downsampling operations enlarge the receptive field; a $96^3$ input produces a final feature map of resolution $6^3$. Global average pooling followed by an MLP yields $\mathbf{g}_0$. Local features describe geometry near candidate locations, while the global descriptor supplies context for the entire shape.

\subsubsection{Stage 1 Base-Sphere Prediction}

Stage~1 places anchors in the voxel grid using a Halton low-discrepancy sequence~\cite{halton1960}, with each anchor $\mathbf{a}_i$ defining a candidate slot. Trilinear sampling of $\mathbf{F}_0$ at the anchor, followed by feature projection, produces a local descriptor $\boldsymbol{\ell}_i$. The prediction head combines local geometry, global context, and slot information:
\begin{equation}
    \mathbf{u}_i=H_0\bigl([
    \mathbf{g}_0,\boldsymbol{\ell}_i,\mathbf{e}_i,
    \mathbf{t}_0,s_i,\tilde{\mathbf{a}}_i]\bigr).
    \label{eq:anchor_slot_prediction}
\end{equation}
Here, $H_0$ is an MLP; $\mathbf{e}_i$ and $\mathbf{t}_0$ are learned slot and stage embeddings; $s_i$ is the normalized slot index; and $\tilde{\mathbf{a}}_i$ is the normalized anchor coordinate. The output $\mathbf{u}_i=(\Delta\mathbf{c}_i,\rho_i,\eta_i)$ contains the center offset, radius parameter, and objectness logit, respectively. These values are converted into sphere parameters as follows:
\begin{equation}
    \begin{aligned}
        \hat{\mathbf{c}}_i^{(0)}
        &=\operatorname{clip}_{[0,R-1]}\bigl(\mathbf{a}_i+
        \delta_c\tanh(\Delta\mathbf{c}_i)\bigr),\\
        \hat{r}_i^{(0)}&=r_{\max}\sigma(\rho_i),
        \qquad \hat{o}_i^{(0)}=\sigma(\eta_i).
    \end{aligned}
    \label{eq:base_sphere_prediction}
\end{equation}
Here, $\operatorname{clip}_{[a,b]}$ denotes componentwise clipping to the specified interval, $\delta_c$ bounds the displacement from the anchor, $r_{\max}$ is the candidate radius limit, and $\sigma$ is the sigmoid function. The implementation additionally applies small numerical bounds to the sigmoid output used for the radius. Larger radius initialization and large-sphere supervision jointly encourage base spheres to represent principal regions with sufficient interior clearance.

\subsubsection{Differentiable Sphere Renderer}

The renderer explicitly exposes the coverage of the base-sphere set to the next stage. At a point $\mathbf{p}$, each sphere contributes a smooth occupancy response, and the responses are combined into a soft union:
\begin{equation}
    \begin{aligned}
        a_i(\mathbf{p})&=\hat{o}_i^{(0)}\,
        \sigma\!\left(\frac{\hat{r}_i^{(0)}-
        \|\mathbf{p}-\hat{\mathbf{c}}_i^{(0)}\|_2}{\tau}\right),\\
        C_0(\mathbf{p})&=1-\prod_{i=1}^{K_0}
        \bigl(1-a_i(\mathbf{p})\bigr).
    \end{aligned}
    \label{eq:coverage_feedback}
\end{equation}
The temperature $\tau$ controls the smoothness of sphere boundaries. Coverage is evaluated on a low-resolution rendering grid; coordinates, radii, and temperature in Eq.~\eqref{eq:coverage_feedback} are all expressed in input-grid units.

Let $A(\mathbf{p})=\sum_i a_i(\mathbf{p})$ denote accumulated soft occupancy, and let $\bar{M}_{\mathrm{in}}$ be the interior mask trilinearly resampled to the rendering grid. The five feedback channels are constructed as follows, with pointwise arguments omitted:
\begin{equation}
    \begin{aligned}
        \mathbf{M}_{\mathrm{fb}}&=\operatorname{Up}([C_0,S,O,U,E]),\\
        S&=\tfrac{1}{2}\operatorname{clip}_{[0,2]}(A),\\
        O&=\tfrac{1}{2}\operatorname{clip}_{[0,2]}(A-1),\\
        U&=\bar{M}_{\mathrm{in}}(1-C_0),\qquad
        E=(1-\bar{M}_{\mathrm{in}})C_0.
    \end{aligned}
    \label{eq:feedback_maps}
\end{equation}
Here, $S$ represents accumulated occupancy, $O$ represents overlap beyond a single coverage contribution, $U$ identifies uncovered interior regions, and $E$ identifies exterior coverage. Clipping and scaling keep the feedback values bounded. The operator $\operatorname{Up}$ trilinearly upsamples the five-channel tensor to the input resolution, exposing missing, redundant, and exterior coverage to the refinement module.

\subsubsection{Stage 2 Correction and Gap Filling}

The second encoder jointly processes the original geometry and the five feedback maps:
\begin{equation}
    (\mathbf{F}_1,\mathbf{g}_1)
    =E_1([\mathbf{V},\mathbf{M}_{\mathrm{fb}}]).
    \label{eq:feedback_encoding}
\end{equation}
The correction head samples local features at base-sphere centers and combines them with current sphere parameters, global context, and slot information to predict bounded increments in center, radius, and objectness logit. Limited blending produces the refined base spheres:
\begin{equation}
    \begin{aligned}
        \hat{\mathbf{c}}_i^{\mathrm{lock}}
        &=\operatorname{clip}_{[0,R-1]}
        (\hat{\mathbf{c}}_i^{(0)}+\beta_c\Delta\mathbf{c}_i^{\mathrm{cor}}),\\
        \hat{r}_i^{\mathrm{lock}}
        &=\operatorname{clip}_{[\varepsilon,R]}
        (\hat{r}_i^{(0)}+\beta_r\Delta r_i^{\mathrm{cor}}),\\
        \hat{o}_i^{\mathrm{lock}}
        &=\sigma(\eta_i+\beta_o\Delta\eta_i^{\mathrm{cor}}).
    \end{aligned}
    \label{eq:bounded_correction}
\end{equation}
The correction head bounds the increments using $\tanh$ and scale factors. The coefficients $\beta_c$, $\beta_r$, and $\beta_o$ lie in $[0,1]$, and $\varepsilon$ is a small positive radius floor. The superscript $\mathrm{lock}$ denotes the base spheres after limited correction. Small blending factors and training-time preservation constraints jointly reduce disruption to existing principal coverage, but do not strictly guarantee monotonically increasing coverage.

The medium head $H_m$ and detail head $H_d$ share the feedback-aware features, predict residual spheres at two scales, and concatenate them with the refined base spheres:
\begin{equation}
    \begin{aligned}
        \hat{\mathcal{S}}_m&=H_m(\mathbf{F}_1,\mathbf{g}_1),\qquad
        \hat{\mathcal{S}}_d=H_d(\mathbf{F}_1,\mathbf{g}_1),\\
        \hat{\mathcal{S}}&=\operatorname{Concat}
        (\hat{\mathcal{S}}_0^{\mathrm{lock}},
         \hat{\mathcal{S}}_m,\hat{\mathcal{S}}_d).
    \end{aligned}
    \label{eq:multiscale_sphere_set}
\end{equation}
Both residual heads use the anchor-based prediction scheme in Eqs.~\eqref{eq:anchor_slot_prediction}--\eqref{eq:base_sphere_prediction}, with different anchor-sequence offsets, radius limits, and initialization scales. Medium spheres address larger residual gaps, while detail spheres use smaller radii to represent slender branches and local structures. Concatenation along the candidate dimension preserves the total capacity $K$, and residual-region supervision directs additional spheres toward areas insufficiently covered by Stage~1.

\subsection{Geometric Training Objective}
\label{subsec:loss_function}

Training uses reference sphere sets generated by the conventional SDF-based greedy IST algorithm together with voxel-geometric constraints. The final set receives the main supervision, while the base set receives auxiliary supervision emphasizing large-scale coverage. Gap filling, base-sphere preservation, feedback-map supervision, and correction regularization are gradually introduced during refinement. The terms are combined with configurable weights and serve the following purposes.

\textbf{Coverage and geometric validity.}
Interior samples encourage volume coverage, exterior samples penalize coverage outside the object, and additional weight is assigned to high-clearance regions and poorly covered samples. Coverage supervision first limits radii using the local distance field, then softly gates activation according to objectness, interior occupancy, and viable radius. Separate validity penalties discourage exterior centers and predicted radii exceeding a conservative fraction of local clearance.

\textbf{Reference-set supervision.}
Sphere sets are unordered~\cite{zaheer2017}, so supervision uses bidirectional nearest geometric matching~\cite{fan2017pointset}. The matching cost combines squared Euclidean center distance and squared radius difference, both normalized by grid resolution, with a weight of 2 on the radius term. Larger reference spheres receive higher weights, predicted spheres are weighted by objectness, and candidates matched to references receive positive objectness supervision. Additional large-sphere supervision helps preserve the principal interior structure.

\textbf{Overlap control and sphere utilization.}
Pairwise constraints emphasize severe penetration and excessive nearest-neighbor overlap. Local tangency constraints and reference-derived contact supervision encourage compact arrangements of neighboring spheres. A soft active-count constraint sets a target based on the reference count and candidate capacity, reducing both insufficient viable activation and excessive activation. Radius-utilization constraints encourage viable spheres to use the available local clearance.

\textbf{Preservation and residual gap filling.}
Preservation constraints discourage excessive shrinking, suppression, and displacement of useful base spheres, and penalize final coverage falling below Stage~1 coverage. Residual supervision emphasizes uncovered interior regions and directs additional centers and coverage toward them. A separate radius-utilization constraint helps medium spheres fill larger gaps. Feedback-map supervision promotes base coverage, and correction regularization limits adjustment magnitude.

\textbf{Optimization schedule.}
The model is optimized with AdamW. The default learning-rate schedule uses warmup followed by cosine decay, with gradient clipping to stabilize updates. Early in training, gradients are detached from feedback maps and base predictions passed to the refiner; later, gradient propagation is restored for joint training. Refinement supervision is introduced gradually, soft-coverage temperatures decrease, and geometric and packing weights are adjusted over training so that the network first establishes principal coverage and then strengthens local refinement.

The full training objective groups the above terms as
\begin{equation}
    \mathcal{L}=\mathcal{J}_f+\lambda_0\mathcal{J}_0
    +s(e)\left(\mathcal{J}_{\mathrm{extra}}+\mathcal{J}_{\mathrm{pres}}
    +\lambda_m\mathcal{J}_{\mathrm{map}}
    +\lambda_\Delta\mathcal{J}_{\Delta}\right),
    \label{eq:objective}
\end{equation}
where $\mathcal{J}_f$ and $\mathcal{J}_0$ collect the weighted geometry and reference penalties for the final and base sets, $\mathcal{J}_{\mathrm{extra}}$ and $\mathcal{J}_{\mathrm{pres}}$ group the gap-filling and preservation terms, $\mathcal{J}_{\mathrm{map}}$ supervises the feedback maps, and $\mathcal{J}_{\Delta}$ regularizes the correction increments. The schedule $s(e)$ ramps the refinement terms with epoch $e$. Implementation defaults and the complete list of supplementary terms are given in Appendix~\ref{app:implementation}.

\subsection{Inference and Validity Projection}
\label{subsec:inference}

Inference consists of one network forward pass, candidate filtering, radius projection, and coordinate conversion. Candidates are initially retained when their objectness reaches the configured threshold and the interpolated interior mask at their centers exceeds 0.35. If their number exceeds the requested output limit, candidates with larger objectness times squared predicted radius are retained preferentially. If no candidate passes this filtering, the implementation falls back to the highest-scoring candidates before applying radius projection and the minimum-radius check.

The radius of each selected candidate is limited by the local distance field:
\begin{equation}
    \hat{r}_i^{\mathrm{legal}}
    =\min\bigl(\hat{r}_i,
    \lambda_{\mathrm{safe}}D(\hat{\mathbf{c}}_i)\bigr),
    \label{eq:radius_projection}
\end{equation}
Here, $D(\hat{\mathbf{c}}_i)$ is obtained by trilinear sampling of the distance field, and $\lambda_{\mathrm{safe}}=0.995$ in the implementation. Spheres below the minimum projected radius are discarded. The remaining spheres are ranked by objectness times squared projected radius, then converted to model coordinates using the original grid dimensions, index origin, and voxel scale. This step enforces the local radius bound supplied by the distance field; actual geometric accuracy still depends on voxel discretization and interpolation error.

Candidate prediction is parallel within each stage, and the pipeline requires no greedy insertion or local search. The generated leaves can be passed directly to subsequent IST hierarchy construction and collision-query procedures.

\section{Experimental Evaluation}
\label{sec:experiments}

This section evaluates the ML-IST leaf-sphere generator in terms of geometric behavior and generation efficiency. Visualizations of complex non-convex models first assess whether the predictions represent principal volumes, slender branches, and variations in local thickness. Generation time is then compared with that of GT-IST at several target sphere counts. The visualizations constitute qualitative geometric evidence, whereas the timing results characterize implementation-level performance under the specified hardware and software configuration. Dynamic usability and physics-engine throughput are evaluated separately in Sections~4 and~5.

\subsection{Experimental Setup}
\label{subsec:experiment_setting}

The platform comprised an Intel Xeon Platinum 8368Q CPU, 48~GB of system memory, and NVIDIA GeForce RTX 4090 GPUs. The software stack included PyTorch, CUDA, and OpenVDB. OpenVDB~\cite{museth2013} constructed the SDF and distance-transform fields from each input mesh, and PyTorch~\cite{paszke2019} implemented and evaluated ML-IST. ML-IST training was distributed across four RTX 4090 GPUs. Generation-time evaluation, however, used a single RTX 4090 for ML-IST inference and a single core of the Xeon 8368Q CPU with SIMD vectorization enabled for the conventional GT-IST greedy search.

The comparison considered GT-IST, the reference leaves produced by conventional greedy maximum-inscribed-sphere search, and the proposed ML-IST.

The dataset comprised 3,780 three-dimensional models. We used 3,600 models as the training set and held out the remaining 180 models as the test set. The test models did not participate in network optimization. The generation-time statistics reported in Section~\ref{subsec:generation_time} were computed over 100 models from the held-out test set.

Parameter $K$ is the target number of leaves in the GT-IST reference, with $K\in\{64,128,256,512\}$. For each value, the conventional method first produced labels containing the corresponding number of spheres, after which a separate ML-IST model was trained for that setting. The four values in Fig.~\ref{fig:generation_time} thus correspond to four independently trained networks. GT-IST time includes greedy maximum-sphere search and leaf output, whereas ML-IST time includes a single-model forward pass and required candidate output. The comparison is restricted to leaf generation under these configurations; subsequent hierarchy construction and collision-query time are excluded.

\subsection{Qualitative Non-Convex Results}

\begin{figure}[tbp]
	\centering
	\includegraphics[width=\linewidth]{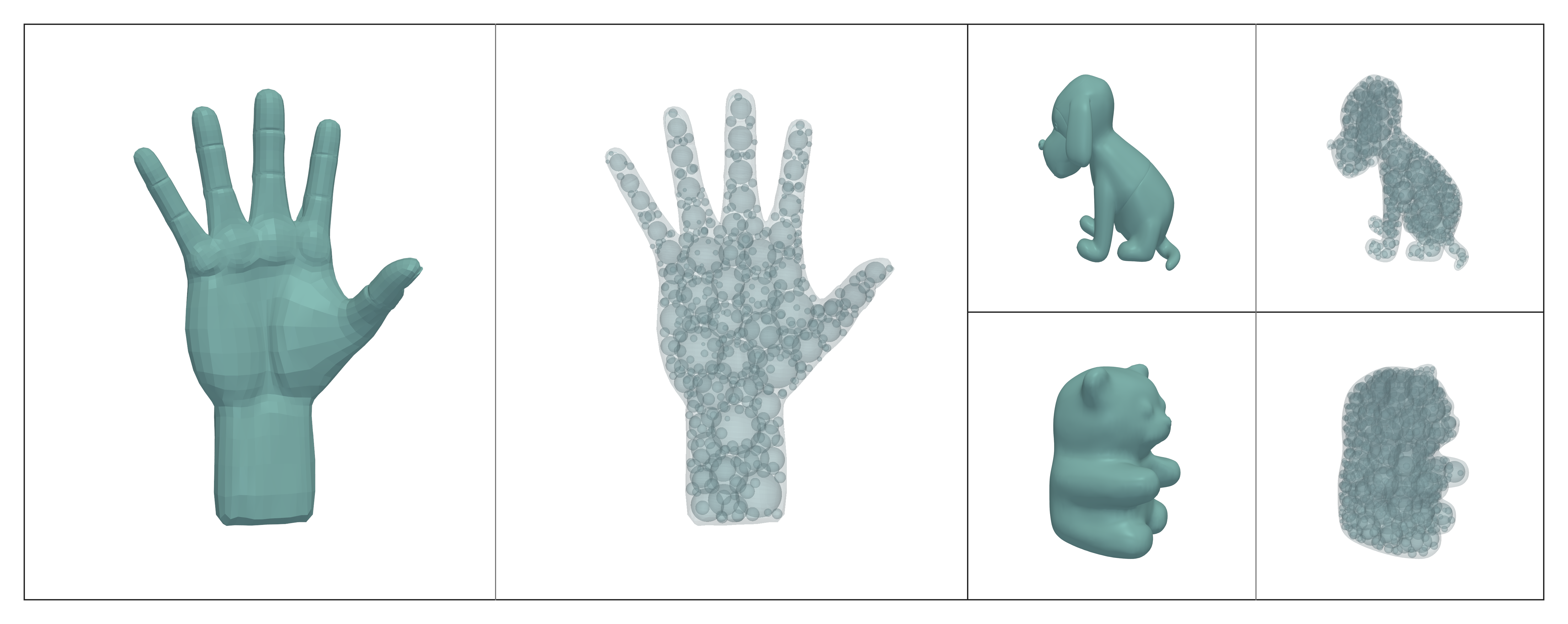}
	\caption{ML-IST inner-sphere sets for complex non-convex models. Each pair shows the input surface mesh and the corresponding ML-IST prediction.}
	\label{fig:visualization_comparison}
\end{figure}

To qualitatively assess geometric plausibility, Fig.~\ref{fig:visualization_comparison} pairs each input surface mesh with its ML-IST prediction for hand-, dog-, and hamster-shaped non-convex models. These examples combine principal volumes of varying thickness with local features such as fingers, limbs, tails, and ears, thereby exposing the network's behavior across geometric scales.

After validity projection, the predicted spheres lie inside the models and their overall distribution follows the input silhouette. Large spheres represent thick principal regions such as the palm, wrist, and animal torso, while smaller spheres extend along fingers, limbs, ears, and changes in contour. Consecutive small spheres follow narrow regions, whereas denser sphere sets occupy the torso of the dog and hamster models, suggesting that the network relates local thickness to admissible sphere scale.

Figure~\ref{fig:visualization_comparison} provides qualitative evidence only: closer visual agreement with the silhouette does not necessarily imply higher collision-detection accuracy. Nevertheless, the projected outputs satisfy the geometric requirements of IST leaves and provide valid inputs to the subsequent experiments.

\subsection{Generation-Time Comparison}
\label{subsec:generation_time}

\begin{figure}[tbp]
	\centering
	\includegraphics[width=\linewidth]{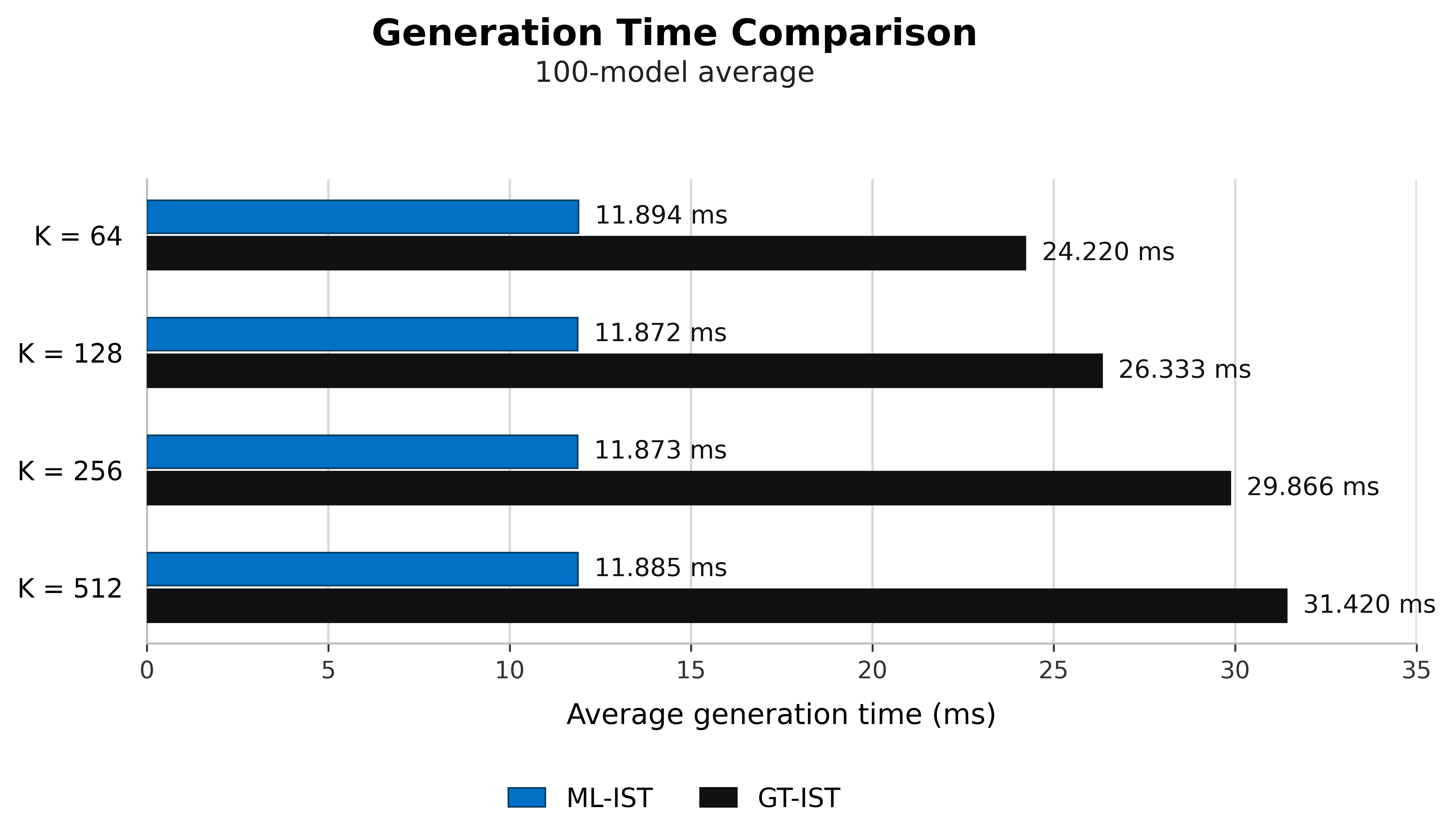}
	\caption{Mean per-model generation time of ML-IST and GT-IST at different target leaf counts $K$, averaged over 100 models from the held-out test set. GT-IST runs on a single core of an Intel Xeon Platinum 8368Q CPU with SIMD vectorization enabled, whereas ML-IST inference runs on one RTX 4090 GPU. Training used four RTX 4090 GPUs, and a separate ML-IST model was trained from GT-IST labels for each $K$.}
	\label{fig:generation_time}
\end{figure}

Figure~\ref{fig:generation_time} compares the mean generation time at $K=64$, 128, 256, and 512. Every ML-IST value was produced by a model trained independently with GT-IST labels at the same $K$. Although model training was distributed across four RTX 4090 GPUs, the reported generation-time measurements use only one RTX 4090 for ML-IST inference. GT-IST runs on a single core of the Intel Xeon Platinum 8368Q CPU with SIMD vectorization enabled. The comparison therefore evaluates one GPU against one CPU core at deployment time.

GT-IST generation time increases as $K$ grows. This trend is consistent with greedy sequential construction: additional target spheres require further maximum-sphere selections and residual-space updates. By contrast, ML-IST evaluates a fixed-depth network and predicts the candidate slots in parallel on a single GPU. Its cost is governed primarily by fixed-resolution 3D feature extraction, with a smaller marginal contribution from increasing the number of output slots.

Under the reported one-CPU and one-GPU inference protocol, ML-IST achieves a lower mean generation time at all four target counts and exhibits less sensitivity to increasing $K$. These results support the use of parallel candidate prediction to reduce the deployment-time cost of sequential greedy leaf construction. When four models are processed concurrently on four RTX 4090 GPUs, the per-model cost amortizes to approximately 3.0~ms. The corresponding quantitative values are reported in Fig.~\ref{fig:generation_time}.

\section{Dynamic Collision Usability}

Section~3 examined geometric appearance and construction time, but neither establishes whether a predicted proxy can participate stably in rigid-body contact. We therefore evaluate ML-IST in a free-fall collision scenario using image sequences, per-step collision-related computation time, and linear-velocity response. The timing measurements exclude generation of both proxies.

\subsection{Free-Fall Collision Scenario}

\begin{figure}[tbp]
	\centering
	\begin{subfigure}[b]{0.22\linewidth}
		\centering
		\includegraphics[width=\linewidth]{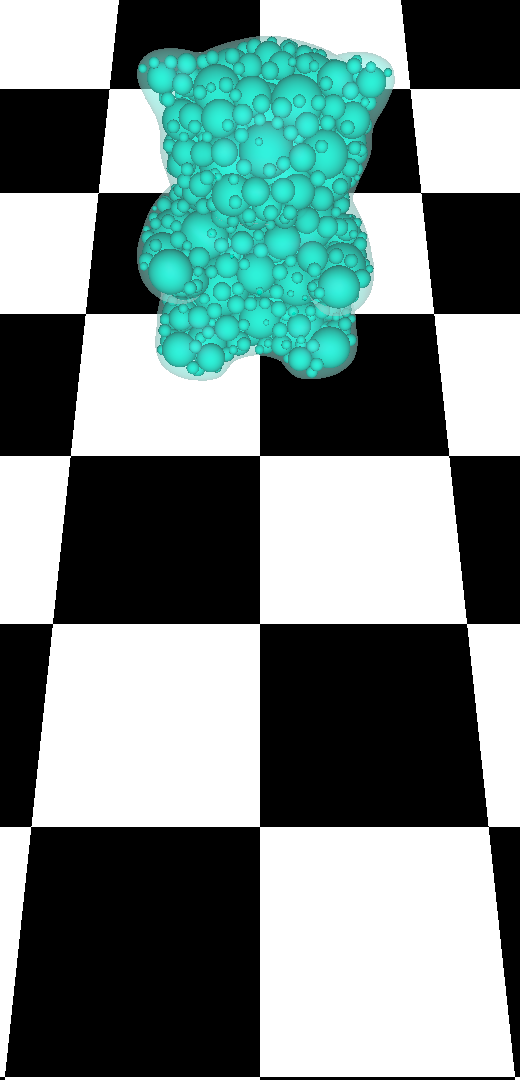}
		\caption{Initial}
	\end{subfigure}\hfill
	\begin{subfigure}[b]{0.22\linewidth}
		\centering
		\includegraphics[width=\linewidth]{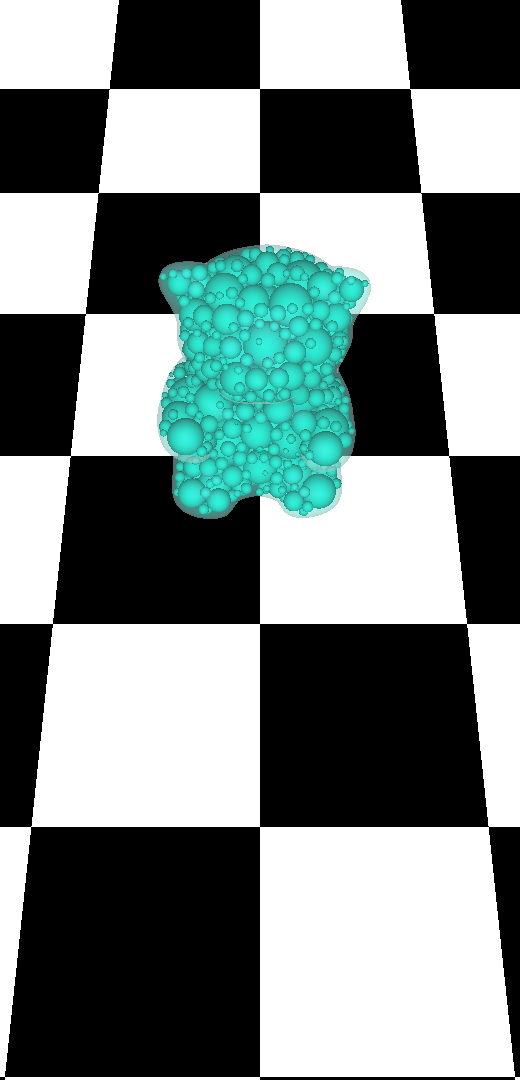}
		\caption{Falling}
	\end{subfigure}\hfill
	\begin{subfigure}[b]{0.22\linewidth}
		\centering
		\includegraphics[width=\linewidth]{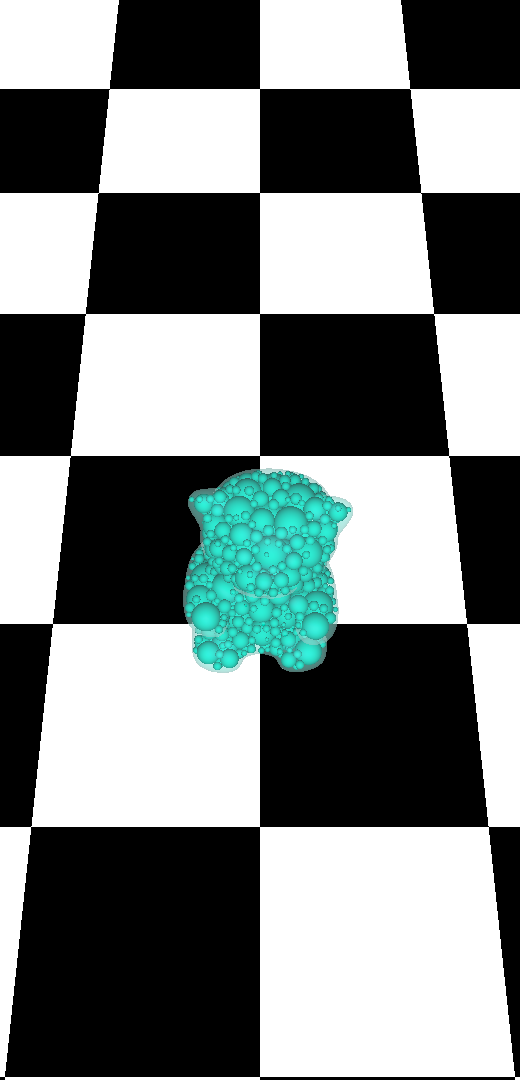}
		\caption{Contact}
	\end{subfigure}\hfill
	\begin{subfigure}[b]{0.22\linewidth}
		\centering
		\includegraphics[width=\linewidth]{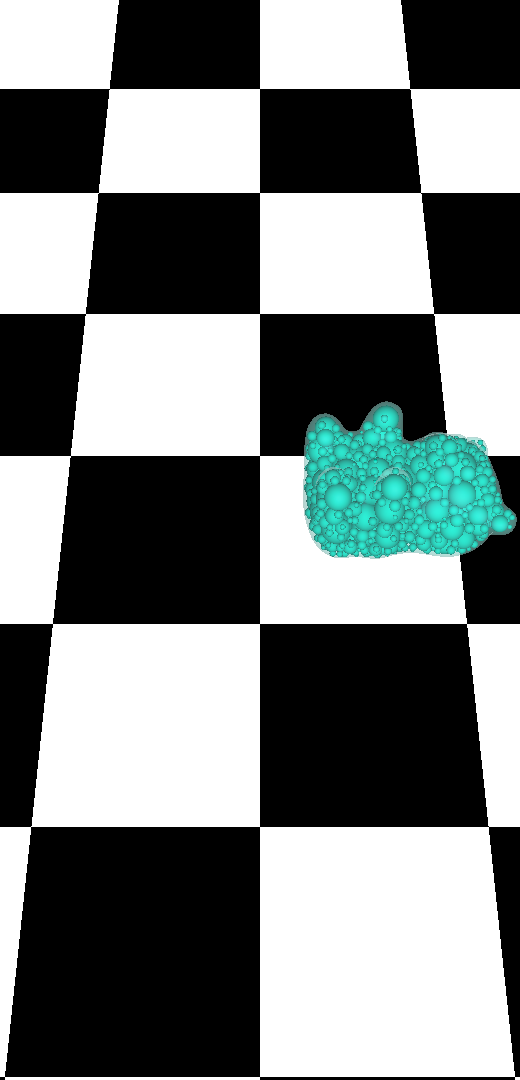}
		\caption{Impact}
	\end{subfigure}
	\caption{Free-fall collision sequence using an ML-IST proxy. The object falls from its initial pose, approaches the plane, and changes orientation after impact.}
	\label{fig:drop_frames}
\end{figure}

The test object has a pronounced non-convex outline and is released above a horizontal plane under gravity. Its ML-IST spheres and visible mesh share the same rigid transform, while the sphere set participates in collision detection and contact response. Figure~\ref{fig:drop_frames} shows four representative instants: the object begins at an elevated position, approaches the plane, and undergoes translation and rotation after contact before settling into a lateral pose. The sequence shows that the predicted set remains spatially consistent with the rigid body and produces an observable contact response.

\subsection{Collision Computation Time}

\begin{figure}[tbp]
	\centering
	\includegraphics[width=\linewidth]{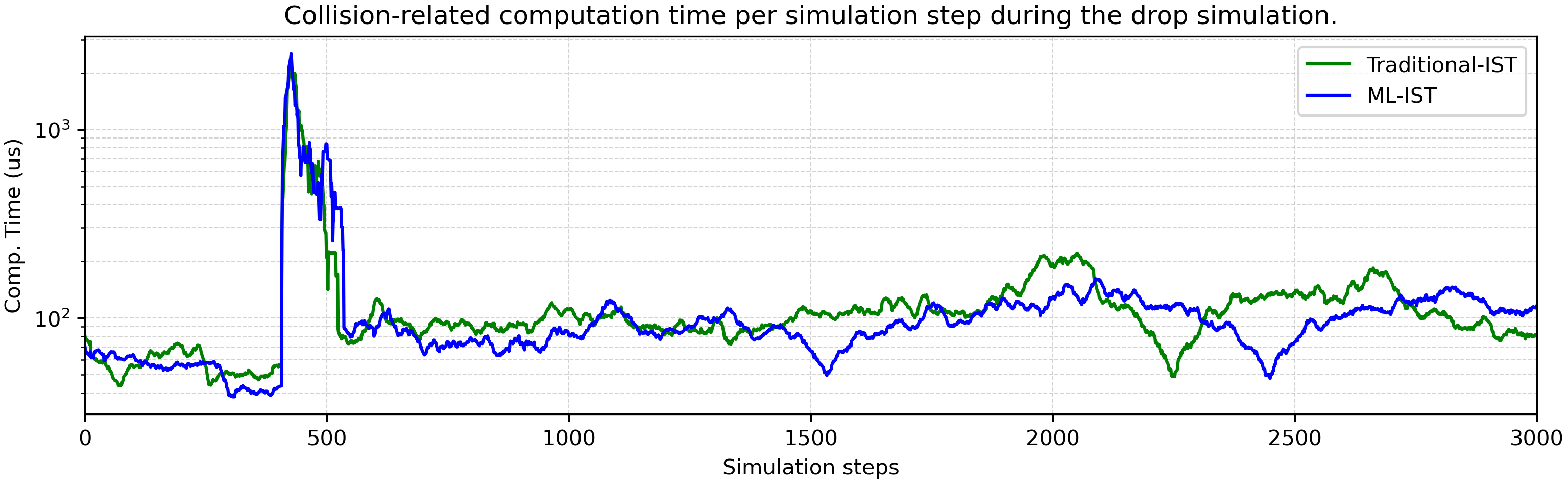}
	\caption{Per-step collision-related computation time of GT-IST and ML-IST in the free-fall simulation. The first 3000 steps are shown using a 20-sample moving average and a logarithmic ordinate.}
	\label{fig:update_time}
\end{figure}

To compare the runtime cost of the two IST proxies after they enter the simulation loop, we record per-step collision-related computation time in the free-fall scenario. Figure~\ref{fig:update_time} shows the first 3000 steps. The measurement aggregates collision detection, contact solving, and related dynamic updates.

Before approximately step~400, the object has not contacted the plane and both curves remain mainly in the tens-of-microseconds range. Near initial contact, both costs rise rapidly to the millisecond range and return to approximately $10^2~\mu\mathrm{s}$ after step~550. The peaks reflect the additional contact-setup and pose-resolution work near impact, in contrast to the motion updates and proximity checks during free fall.

After contact, the curves remain within the same order of magnitude and alternate in relative rank over time; neither proxy is uniformly faster throughout the simulation. Replacing the conventional spheres with ML-IST predictions therefore introduces no order-of-magnitude overhead in per-step dynamics. Together with Section~\ref{sec:experiments}, this result indicates that the primary efficiency gain arises during proxy generation, while the generated set retains a dynamic-update cost comparable to GT-IST.

\subsection{Linear-Velocity Response}

\begin{figure}[tbp]
	\centering
	\includegraphics[width=\linewidth]{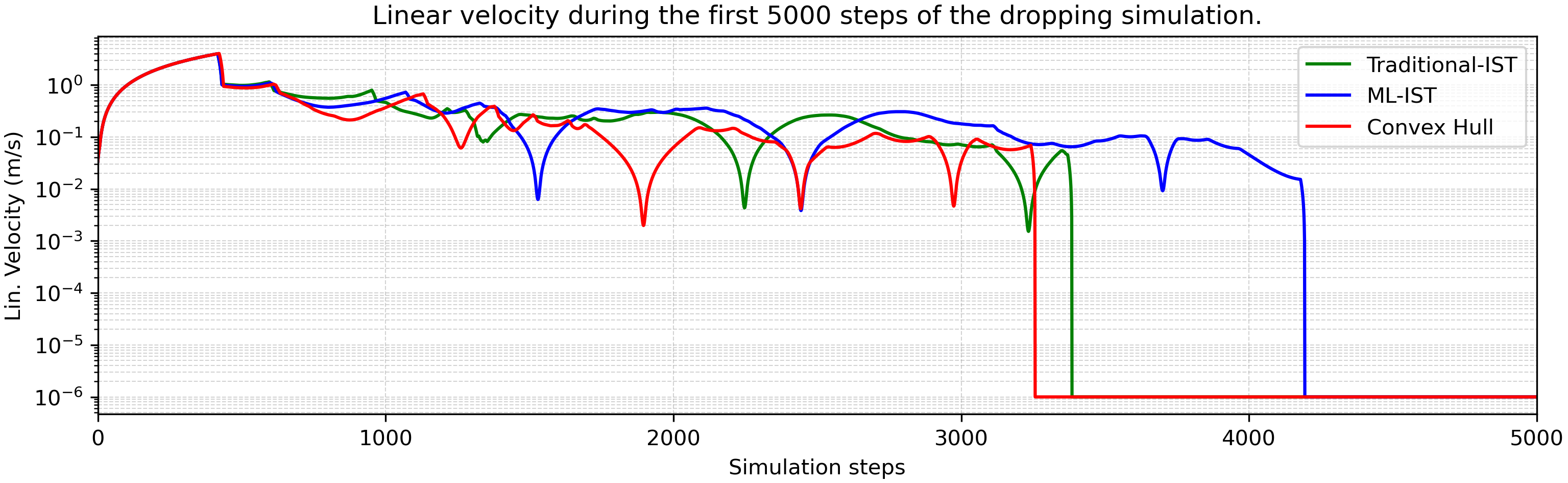}
	\caption{Linear velocity of GT-IST, ML-IST, and convex-hull proxies over the first 5000 steps of the free-fall collision simulation. The ordinate is logarithmic.}
	\label{fig:linear_velocity}
\end{figure}

We additionally record linear velocity over the first 5000 steps for GT-IST, ML-IST, and a convex-hull proxy (Fig.~\ref{fig:linear_velocity}). The plateau at $10^{-6}\,\mathrm{m/s}$ is the plotting floor and is treated as a stationary state.

The three curves nearly coincide before initial contact: gravity accelerates the object from near zero to approximately $4\,\mathrm{m/s}$. After impact, the speed rapidly falls to about $1\,\mathrm{m/s}$ and then undergoes several damped oscillations, reproducing the basic sequence of free fall, contact, and collision response. Their trajectories diverge after contact because the geometric approximation changes the contact distribution, torque, and subsequent pose, which in turn shifts the timing of rebounds and decay.

The convex-hull proxy reaches the stationary floor at approximately step~3250, GT-IST at step~3400, and ML-IST at step~4200. Both IST proxies exhibit more post-impact oscillations than the convex hull, but their decay trajectories are not identical because their local centers and radii differ. The experiment demonstrates that the ML-IST proxy produces a stable and complete dynamic response.

\section{Runtime System Evaluation}
\label{sec:online_realtime}

Sections~3 and~4 evaluated leaf-sphere generation and dynamic usability, respectively. A broader implication of replacing IST search with GPU inference is the possibility of constructing a detailed inner-sphere representation while a physics engine is running. This section motivates runtime generation of collision structures and integrates ML-IST into production physics engines for a system-level comparison with commonly used detailed collision representations.

\subsection{Motivation for Runtime Collision Cooking}
\label{subsec:online_cooking}

In a physics engine, a rendering mesh is generally not an efficient collision representation. The raw geometry is therefore converted into an internal form through a process commonly called cooking; PhysX convex-mesh cooking and Unity mesh cooking, for example, build spatial-search structures for physics queries and contact generation~\cite{nvidia_physx_geometry,unity_mesh_collider}.

Two application classes should therefore be distinguished. In offline pregeneration, scene geometry is known before execution and collision data can be cached during import or loading, leaving only retrieval at runtime; NVIDIA Omniverse similarly notes that cached cooking data avoids repeated computation~\cite{omniphysx_sim_control}.

In runtime generation, by contrast, geometry is created or modified only after execution begins, so no pregenerated collision data are available---for example, when new CAD models are created for immediate assembly, digital-twin geometry is updated from sensor streams, or content is produced procedurally at runtime.

Here, construction time is part of runtime performance rather than merely a preprocessing metric: tens or hundreds of milliseconds of cooking can block the main or physics thread or delay a new object's entry into the simulation. Unity documents an initial CPU spike from runtime cooking~\cite{unity_collider_performance}, and Omniverse states that the main thread waits when required cooking data are not cached at simulation start~\cite{omniphysx_colliders}.

This is especially constraining for real-time physics, which engines commonly update at a fixed step such as Unity's default 50~Hz~\cite{unity_fixed_timestep}: if collision construction exceeds the frame budget, later frames must execute multiple physics steps to catch up. Real-time simulation therefore requires collision generation close to a single-frame budget.

This setting motivates ML-IST: conventional ISTs remain suitable for offline generation, but for complex shapes that appear at runtime, ML-IST casts construction as GPU inference and can provide a detailed inner-sphere representation at near-real-time latency.

\subsection{Physics-Engine Benchmark Setup}
\label{subsec:engine_test_setup}

ML-IST was integrated into Jolt Physics~\cite{joltphysics_repo} and Bullet and compared with three commonly used detailed collision representations:

\begin{itemize}[leftmargin=2em]
	\item \textbf{ML-IST}: the proposed learning-based inner sphere tree;
	\item \textbf{GT-IST}: a detailed inner sphere tree generated by conventional greedy search;
	\item \textbf{GImpact}: Bullet's triangle-mesh collision structure~\cite{bullet_gimpact_repo}; and
	\item \textbf{Convex Hull with Convex Decomposition}: a collision representation based on convex decomposition and convex hulls~\cite{mamou2009}.
\end{itemize}

The alpha puzzle was selected as the benchmark. Its two tightly fitting components require a detailed collision representation and exercise complex geometric fit, narrow-clearance contact, and resistance to visible interpenetration. For a task-level comparison, all four representations were configured so that the components could be inserted without visible interference or tunneling.

This benchmark used one NVIDIA GeForce RTX 4090 GPU and an Intel Xeon Platinum 8368Q CPU. An isolated inference test required approximately 12~ms to generate one IST model on the RTX 4090. The integrated engine additionally incurs GPU-CPU transfer and synchronization costs; under the complete benchmark, ML-IST operated at approximately 55~Hz, close to the 50--60~Hz range commonly used for real-time simulation.

\subsection{Results and Analysis}
\label{subsec:engine_test_result}

\begin{figure}[tbp]
	\centering
	\includegraphics[width=\linewidth]{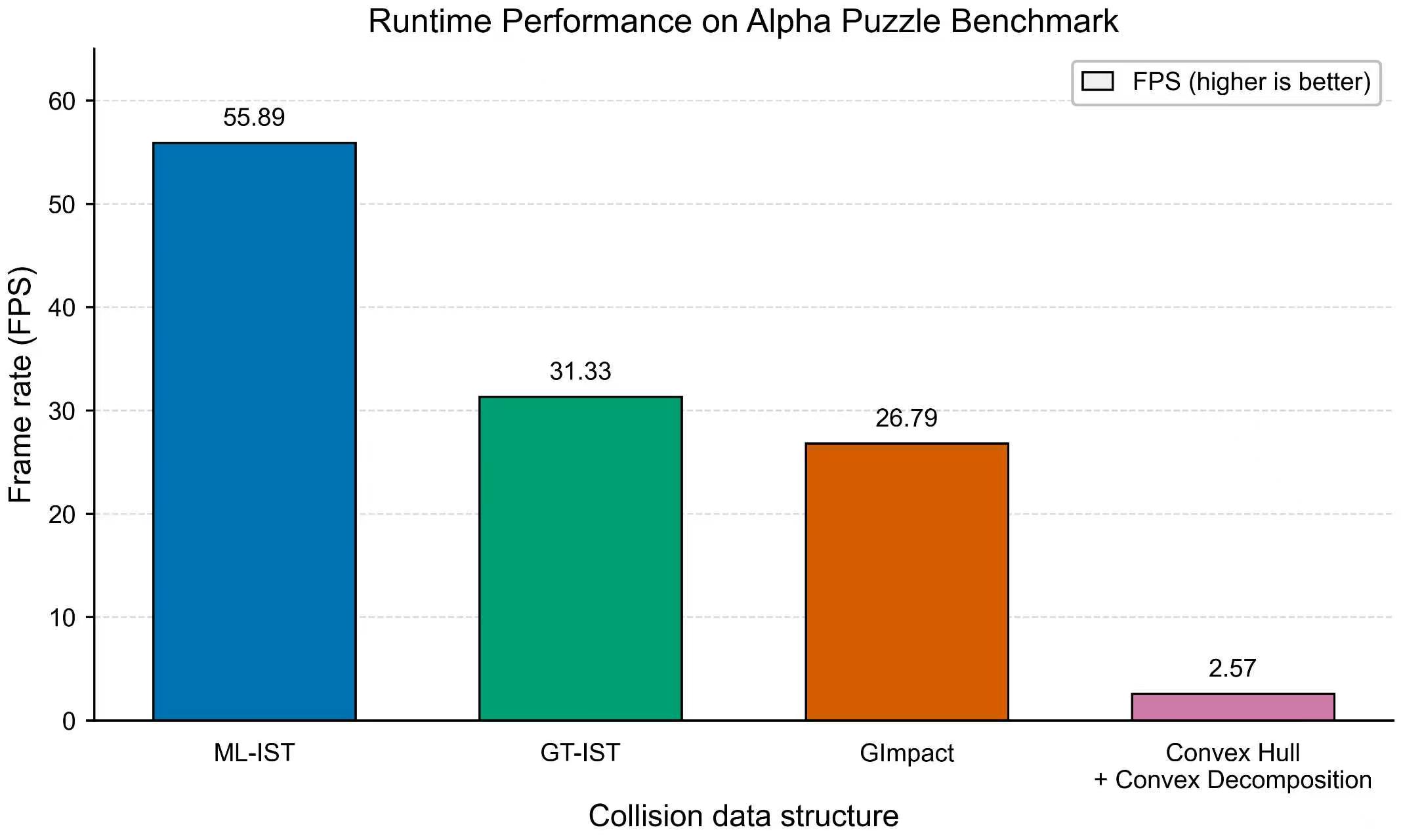}
	\caption{Runtime frame rate of four collision representations in the alpha-puzzle benchmark. ML-IST reaches 55.89~FPS, compared with 31.33~FPS for GT-IST, 26.79~FPS for GImpact, and 2.57~FPS for convex hulls with convex decomposition.}
	\label{fig:fps_compare}
\end{figure}

Figure~\ref{fig:fps_compare} reports the runtime frame rates of the four collision representations in the alpha-puzzle scene.

On the reported platform and scene, ML-IST achieves the highest measured frame rate. Its 55.89~FPS is 78.4\% higher than the 31.33~FPS of GT-IST. It also exceeds GImpact at 26.79~FPS and convex decomposition at 2.57~FPS; the latter is well below typical interactive rates in this benchmark. These are system-level measurements and jointly reflect representation generation, integration, and physics-engine execution rather than isolated network latency alone.

The result is relevant to applications that create complex objects dynamically. Conventional pipelines often trade construction latency against proxy detail: a detailed representation is expensive to build, whereas a rapidly generated coarse proxy can alter contact behavior. By casting leaf generation as GPU inference, ML-IST provides a practical path toward runtime IST construction while retaining the task-level geometric fit required by the benchmark.

\subsection{Summary}
\label{subsec:engine_test_summary}

This section explained why collision-structure generation contributes directly to runtime performance in industrial simulation, digital twins, interactive games, and robotics. ML-IST was then integrated into Jolt Physics and Bullet and compared with GT-IST, GImpact, and convex hulls with convex decomposition in the alpha-puzzle benchmark.

On the RTX 4090 and Xeon 8368Q platform, ML-IST reached 55.89~FPS, compared with 31.33~FPS for GT-IST, 26.79~FPS for GImpact, and 2.57~FPS for convex decomposition. Under this task and implementation, ML-IST combines faster IST generation with interactive physics-engine throughput. Broader evaluation should include more models and analyze topology, voxel resolution, SDF-query accuracy, timing variance, and GPU-CPU transfer strategies.

\section{Conclusion}

This paper introduced ML-IST, a learning-based leaf-sphere generator that reduces the construction cost of conventional inner sphere trees by reformulating sequential greedy placement as fixed-capacity set prediction. A multichannel voxel representation, 3D residual encoder, Halton-anchor slots, and differentiable coverage renderer support coarse-to-fine prediction of candidate leaves, which are projected onto the geometrically valid domain before conventional IST hierarchy construction.

Across target leaf counts $K=64$, 128, 256, and 512, ML-IST reduced leaf-generation time by a factor of 2.0--2.6 compared with GT-IST on a single CPU core while showing less sensitivity to increasing $K$. In a free-fall scenario, the predicted spheres exhibited collision-update costs of the same order of magnitude as GT-IST and produced a stable dynamic response. Integrated into Jolt Physics and Bullet, ML-IST reached 55.89~FPS in the alpha-puzzle benchmark, exceeding GT-IST, GImpact, and convex decomposition.

ML-IST therefore converts geometric optimization for IST leaf generation into data-driven candidate-set prediction, offering a learned proxy-generation strategy for rapid IST construction. Future work includes broader evaluation across model topologies and voxel resolutions, analysis of timing variance and GPU-CPU transfer strategies, and runtime preprocessing for collision detection, physics simulation, and interactive virtual environments.

\section*{Code Availability}

The official implementation of ML-IST, including mesh preprocessing, ground-truth generation, training, prediction, and visualization, is publicly available at the following repository:
\begin{center}
	\url{https://github.com/HuangJianNeng/ML-IST}
\end{center}

\appendix
\section{Implementation Details}
\label{app:implementation}
\subsection{Defaults and Training Schedule}
Table~\ref{tab:defaults} lists defaults of the supplied implementation. In particular, its default capacity of 700 is not the capacity used for every reported experiment; actual commands and checkpoint configurations can override these values.
\begin{table}[tbp]
\centering
\caption{Reference code defaults; not asserted as the settings of the reported experimental runs.}
\label{tab:defaults}
\small
\begin{tabular}{@{}p{0.54\linewidth}p{0.37\linewidth}@{}}
\toprule
Setting & Default\\
\midrule
Input resolution & $96\times96\times96$\\
Candidate capacity & 700\\
Base / medium capacity fraction & 0.30 / 0.32\\
Encoder base channels & 24\\
Global feature / slot embedding & 384 / 96\\
Optimizer & Adam\\
Learning rate / weight decay & $10^{-4}$ / $7\times10^{-5}$\\
Adam moments & 0.9 / 0.95\\
Sampled points per object & 16384\\
Interior sampling fraction & 0.86\\
Mixed precision in training & Disabled\\
Gradient norm clipping & 1.0\\
Objectness gate threshold / width & 0.40 / 0.10\\
Effective minimum clearance & 0.75\\
Effective radius safety factor & 0.995\\
Radius validity factor & 0.98\\
Correction blends $(\beta_c,\beta_r,\beta_o)$ & 0.42, 0.36, 0.42\\
Inference occupancy threshold & 0.35\\
Inference output limit & target capacity $K$\\
Minimum projected radius & 0.30\\
Hierarchy branching factor & 4\\
\bottomrule
\end{tabular}
\end{table}

Refinement penalties ramp from 6\% to 38\% of the training horizon. Feedback and base predictions are detached from the second stage until approximately 62\% of the horizon, after which joint gradients are enabled; the preservation reference and residual weights remain explicitly detached. Occupancy temperatures, active budgets, and packing weights vary during training, and the best checkpoint is selected by a composite score of training-epoch geometry metrics.

\subsection{Supplementary Loss Components}
The final and Stage~1 objectives include different weights and subsets of geometry penalties. Beyond the coverage, matching, overlap, and active-count mechanisms summarized in Section~\ref{subsec:loss_function}, these include hard-point and focal coverage, point-to-sphere margins, center-distribution coverage, gap-weighted center and margin errors, count underflow and overflow, and radius-utilization terms for viable and large spheres. Packing is supplemented by worst-neighbor overlap, voxel-scale deep-overlap guards, smooth pair barriers, and a contact teacher derived from matched reference spheres. Binary-objectness and radius regularization are also used.

The residual group adds center and margin terms, a residual active-budget term, and a clearance-guided medium-radius term. The preservation group adds penalties on negative radius/objectness corrections for useful base spheres, hard-point coverage reduction, and geometric-margin reduction. Feedback supervision penalizes uncovered interior and low soft interior coverage. Increment regularization penalizes squared center, radius, and logit changes.

For self-containedness, the most influential of these terms are briefly characterized here; the full configuration defines the remainder. \emph{Hard-point coverage} averages the negative log soft coverage over the worst-covered interior samples, whereas \emph{focal coverage} reweights the coverage loss by $(1-C)^2$ so that gradients persist on regions that are only partially covered. \emph{Point-to-sphere margins} require, for each interior sample, the containment margin $r-\|\mathbf{p}-\mathbf{c}\|$ of its best viable sphere to reach a target that grows with the local clearance, and \emph{center-distribution coverage} requires every interior sample to lie within a clearance-dependent band of an active sphere center. Their \emph{gap-weighted} variants reweight these errors by the residual coverage deficit $(1-C)^{1.5}$, directing refinement toward regions that the base spheres still leave uncovered. \emph{Worst-neighbor overlap} tracks, per sphere, the deepest overlap with any single neighbor so that one bad pair is not diluted by all-pair averaging; \emph{voxel-scale deep-overlap guards} measure penetration depth in absolute voxel units rather than relative to radius; and \emph{smooth pair barriers} apply a softplus barrier that penalizes pairs already as they approach contact. The \emph{contact teacher} transfers the near-tangent gap structure of matched reference sphere pairs to the corresponding predicted pairs. \emph{Deep-large point coverage} emphasizes high-clearance interior points, demanding that they be covered by large spheres.

These mechanisms are implemented in \texttt{sphere\_losses}, \texttt{extra\_residual\_losses}, and \texttt{preservation\_losses}, as well as in \texttt{map\_supervision\_loss} and \texttt{compute\_losses}. Their weights and adaptive schedules are configuration-dependent. The main text describes the principal geometric operations and the overall grouping in Eq.~\eqref{eq:objective}; a reproducible run requires the saved configuration and the corresponding code version, including these supplementary terms.

\end{document}